\PassOptionsToPackage{table}{xcolor}
\documentclass[runningheads]{llncs}
\usepackage[T1]{fontenc}
\usepackage{graphicx}
\usepackage{booktabs}
\usepackage{amsmath,amssymb}
\usepackage{multirow}
\usepackage{xcolor}
\usepackage{colortbl}
\usepackage{siunitx}
\usepackage{xspace}
\usepackage[colorlinks,citecolor=blue!70!black,urlcolor=blue!70!black,linkcolor=blue!70!black]{hyperref}
\usepackage[capitalize,noabbrev]{cleveref}

\newcommand{\DDS}{DermDepth\textsubscript{S}}
\newcommand{\new}[1]{#1}

\newcommand\blfootnote[1]{\begingroup\renewcommand\thefootnote{}\footnotetext{#1}\endgroup}
\begin{document}
\title{DermDepth: Toward Monocular Metric Scale 3D Reconstruction Models for Dermatology}
\titlerunning{DermDepth}

\author{Héctor Carrión\inst{1} \and Narges Norouzi\inst{2}}
\authorrunning{Carrión et al.}
\institute{University of California, Santa Cruz, USA \and
University of California, Berkeley, USA \\
\email{hcarrion@ucsc.edu, norouzi@berkeley.edu}}

\maketitle
\blfootnote{Author's accepted manuscript. Accepted at MICCAI 2026 (Springer LNCS). The final authenticated version will be linked here upon publication.}
\begin{abstract}
Dermatological practice routinely involves measuring and tracking lesion size, morphology and texture, as critical components of wound or skin cancer screening, monitoring and diagnosis. To accomplish this task, practitioners often image the skin surface with commonly available off-the-shelf camera sensors. This has led to an overwhelming research focus on 2D methods while these objectives naturally benefit from 3D information. In this paper, we demonstrate that dense monocular 3D reconstructions, metric scale measurements and rich surface normal texture estimates are achievable for both dermoscopic and macroscopic cases without the need for additional hardware or multiple captures. We present \textbf{DermDepth}, the first single-view metric scale 3D model for the dermatological domain and \textbf{D-Synth}, the first synthetic dermoscopic dataset with pixel-perfect 3D information. Our experiments show training DermDepth on D-Synth corrects metric scale error from over 16$\times$ to under 1.1$\times$ for real dermoscopic data, while preserving geometric quality and increasing texture richness. Fine-tuning on a small amount of real clinical samples generalizes our method across three real-world benchmarks spanning the few mm to hundred cm range, diverse skin-tones, chronic wound cases and produces measurements broadly consistent with disease size reported in medical literature. All code, data and models are available at \url{https://github.com/hectorcarrion/dermdepth}.

\end{abstract}

\section{Introduction}
\label{sec:intro}
Skin lesions and chronic wound conditions impact the lives of millions of patients all over the world with an estimated 3 billion people lacking access to adequate dermatological care, especially in poor communities \cite{coustasse_2019_use}. Non-melanoma skin cancer is the most common malignant tumor with its incidence increasing year over year and melanoma cancers are highly accretive, meaning morphology undergoes progressive growth \cite{wang-2024-cancer}. Early detection significantly increases survival rates \cite{balch-2009}, emphasizing the scale and importance of active Asymmetry, Border, Color, Diameter and Evolution (ABCDE measurements) in dermatology \cite{blum-2003}; features that Artificial Intelligence (AI) algorithms could facilitate tracking.

\begin{figure}[t]
    \centering
    \includegraphics[width=\textwidth]{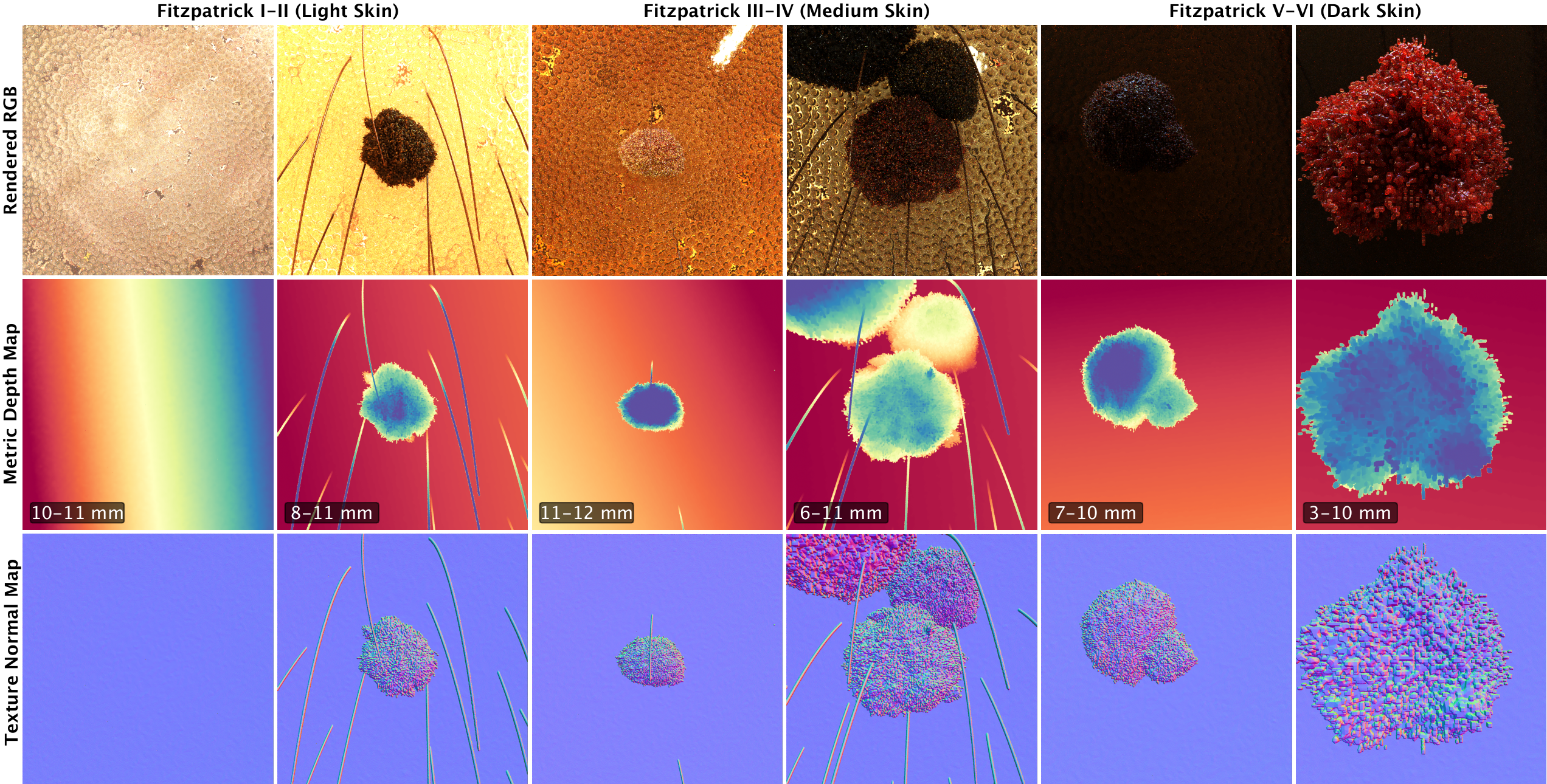}
    \caption{\textbf{D-Synth Dataset.}
    Rendered skin-lesion synthetics from our D-Synth pipeline with dense pixel-perfect 3D information and texture normal maps.
    }
    \label{fig:D-Synth}
\end{figure}

As such, dermatology is often a measurement problem: clinicians screen, triage, and monitor lesions or wounds by tracking changes in size, border irregularity, surface morphology, and texture over time \cite{argenziano-1998}. These characteristics are inherently 3D, yet the standard information capture at point of care remains 2D imaging, likely due to cost and inaccessibility of 3D systems. Critically, these measurements require scale consistency, ideally in metric space, thus 2D images are sometimes paired with rulers. However, clinically valuable information such as elevation, indentation or surface roughness are missed by simple width estimates.

The primary bottleneck toward training and evaluating 3D AI methods in dermatology is the difficulty of collecting suitable datasets \cite{xie-2025}. Purpose-built devices have enabled limited 3D dermatological capture: SKINL2~\cite{de-faria-2019} provides dermoscopic depth via plenoptic camera and WoundsDB~\cite{juszczyk-2020} captures clinical wounds via time-of-flight, demonstrating significant downstream benefits: increased classification accuracy ($\uparrow$10\%), sensitivity ($\uparrow$29\%) \cite{pereira-2021}, and $3\times$ more accurate area estimation \cite{juszczyk-2020}, but require special hardware. Existing monocular approaches are limited to coarse elevation classification \cite{abhishek2024lesion}, relative-scale defocusing \cite{parekh-2024}, or multi-view private-data methods \cite{zhang-2025}. Synthetic dermatological data~\cite{sinha2024dermsynth3d,kim-2024} has been explored for 2D tasks but not for metric 3D depth.

Meanwhile, natural-image monocular depth estimation has advanced rapidly by leveraging large vision transformers~\cite{oquab2023dinov2}. Recent methods claim metric depth output: MoGe-2~\cite{wang2025moge2} and MapAnything~\cite{keetha2026mapanything} predict scale via decoupled heads, Depth Anything v3~\cite{depthanything3} uses canonical camera transformations, and Pixel-Perfect Depth~\cite{xu2025pixel} refines depth via diffusion generation. Whether these models generalize to dermatological images remains underexplored. In order to address these challenges, we present the following contributions:

\begin{enumerate}
    \item \textbf{D-Synth}, a novel rendered synthetic dataset leveraging knowledge informed lesion growth models \cite{kim-2024}, contributing pixel-perfect 3D depth in metric scale, surface normals and camera intrinsics, as shown in \cref{fig:D-Synth}.

    \item \textbf{DermDepth}, a model and fine-tuning approach for affine-invariant 3D prediction methods, correcting metric scale error across benchmarks by up to a factor of 14$\times$, while improving surface normal detail and performing equitably across Fitzpatrick skin tones (disparity: 10.9~$\rightarrow$~1.02).

    \item \textbf{Systematic evaluation} of metric scale accuracy on dermatological images across 4 foundation models, 3 evaluation datasets, and 2 imaging modalities.

    \item \textbf{Downstream validation} of metric accuracy for 3D lesion measurement (area, width, volume) from single photographs, validated against ruler ground truth and literature review.
\end{enumerate}

\section{Method}
\label{sec:method}

\subsection{Training and Evaluation Datasets}
\label{sec:datasets}

\begin{table}[t]
\centering
\caption{\textbf{DermDepth Datasets.} SKINL2 is stratified by version, WoundsDB by patient, and DDI by Fitzpatrick skin tone.}
\label{tab:datasets}
\small
\setlength\tabcolsep{4pt}
\begin{tabular}{@{}l l r r r l@{}}
\toprule
Dataset & Modality & Total & Train & Test & 3D GT source \\
\midrule
D-Synth  & Synth.\ dermoscopy & 4{,}000 & 3{,}000 & 1{,}000   & Rendered depth \\
SKINL2   & Dermoscopy         &     375 &     263 & 112   & Plenoptic camera \\
WoundsDB & Clinical wound     &      77 &      50 &  27   & ToF sensor \\
DDI      & Clinical + ruler   &      47 &      33 &  14   & Ruler annotation \\
\bottomrule
\end{tabular}
\end{table}

We extend \cite{kim-2024}, which renders dermoscopic-style samples following biologically motivated lesion growth patterns. Originally, the method was designed to export 2D images and binary masks; we additionally contribute functionality to export per-pixel depth in metric scale, camera intrinsics and normal maps, while adding camera angles and multiple lesions. We synthesize 4{,}000 samples across 131 melanin levels and 240 lesion morphologies at 12--20\,mm distance with $75^\circ$ field of view. We call this new dataset D-Synth (\cref{fig:D-Synth}). \new{We retain realism features including layered melanosome, blood and lipid models in epidermis, dermis and hypodermis, alongside physics-based light scatter across wavelengths and skin tones, however, we note a photorealism gap which real data can alleviate}.

\label{sec:realdata}
To bridge this sim-to-real gap and evaluate across clinical distances, we process SKINL2 and WoundsDB for fine-tuning and evaluation. SKINL2 plenoptic depth maps exhibit local planar noise~\cite{lourenco-2022} and WoundsDB time-of-flight maps are sparse, low-resolution with spatial offset from the RGB sensor. These issues may contribute noise to geometric evaluation but do not impact metric scale.

\label{sec:ddi}
Diverse Dermatology Images (DDI) \cite{daneshjou_2022_disparities} is a fairness-focused 2D dataset akin to smartphone capture during first-pass clinical examination, sitting between SKINL2 and WoundsDB in terms of sampling distance. This dataset is collected without 3D information, however, we find 47 samples include a fully visible medical ruler with known dimensions \cite{delasco} and matching segmentation masks \cite{carrion-2023}. This allows us to estimate the metric scale of the scene and evaluate fairness across skin-tones. \cref{tab:datasets} summarises all datasets and their train/test splits.

\subsection{Model Architecture and Training Strategy}

\label{sec:architecture}
MoGe-2 and Pixel-Perfect Depth \cite{wang2025moge2,xu2025pixel} predict affine-invariant depth maps which can then be scaled to metric space. This metric scaling factor is predicted via dedicated output heads by both MoGe-2 and MapAnything \cite{keetha2026mapanything} with the former also able to predict normal maps via dedicated convolutional head. Given that real dermatology datasources may have accurate scale but noisy depth (and thus normals), we elect MoGe-2 as our base architecture due to its design flexibility.


\label{sec:training}
When training on D-Synth we freeze the backbone encoder and fine-tune the scale and normal output heads. The scale head is composed of a 3-layer Multi-Layer Perceptron (MLP) that takes the CLS token from the base encoder as input and outputs a single scalar in log-space. The scale factor is then multiplied into the predicted affine-invariant point cloud to convert from relative to metric units, the loss is thus defined as:

\begin{equation}
    \mathcal{L}_{\text{scale}} = \left(\log \hat{s} - \log s^*\right)^2
\end{equation}

\noindent Where $\hat{s}$ is the predicted scale factor and $s^*$ is the ground-truth value. This formulation makes the loss symmetric with respect to over- or under-estimation, which is useful since monocular scale can be ambiguous. The normal output head is composed of a ConvStack \cite{wang2025moge2} decoder similar to the 3D head, inputting feature maps and outputting a 3-channel per-pixel normal vector that is L2-normalized to unit length. The normal loss formula is:


\begin{equation}
\mathcal{L}_{\text{normal}} = \frac{1}{HW} \sum_{i} \angle(\hat{\mathbf{n}}_i, \mathbf{n}_i^*)^2
\end{equation}

\noindent Where $H, W$ are image dimensions, $\hat{\mathbf{n}}_i$ and $\mathbf{n}_i^*$ are the predicted and ground-truth unit surface normals at pixel $i$, and $\angle(\cdot,\cdot)$ denotes the angle in radians between two vectors. When training on SKINL2, WoundsDB or DDI, given their noisy depth and normal maps, we choose to freeze the normal output head as well.

We split datasets 70\% for training and 30\% for testing stratified across versions for SKINL2, patient cases for WoundsDB (so that no patient appears in both splits) and skin-tones for DDI. We train for 1{,}000 steps with learning rate $10^{-4}$, AdamW optimiser, and batch size 4 on a single A100 GPU.
Exponential moving average (EMA, decay 0.999) is applied to all parameters; we find EMA essential for stable convergence, as raw checkpoints exhibit oscillation and overcorrection. The same protocol is used for real-data training stages, with learning rate reduced to $5 \times 10^{-5}$ and 2{,}000 steps per stage.

\subsection{DDI Fairness and Lesion Measurements}
\label{sec:downstream}
We estimate lesion width, area and volume for held-out DDI samples with visible rulers for validation. Given predicted metric depth and estimated camera intrinsics, we back-project to a 3D point cloud and compute: surface area via cross-product triangle integration, lesion width as the major PCA axis, and volume as the integral between the lesion surface and a fitted boundary plane.

\section{Results}
\label{sec:results}

\begin{table}[t]
\centering
\caption{\textbf{Scale Results.} Foundation models systematically miss scale on dermatological data despite strong geometry (SI-$\delta_1$). DermDepth fine-tunes only 2.1\,M parameters (0.6\,\%), with significant scale correction on all benchmarks. Scale ratio target is 1.0$\times$; DDI ratio target is 1.0$\times$. \DDS{}: synthetic data only. Eval on held-out test sets.}
\label{tab:main}
\small
\setlength\tabcolsep{4.5pt}
\begin{tabular}{@{}l c cc cc c@{}}
\toprule
\multirow{2}{*}{Method} & \multirow{2}{*}{Params} & \multicolumn{2}{c}{SKINL2 (dermoscopy)} & \multicolumn{2}{c}{WoundsDB (clinical)} & DDI \\
\cmidrule(lr){3-4} \cmidrule(lr){5-6} \cmidrule(lr){7-7}
& & Scale~$r$ & SI-$\delta_1$\,(\%) & Scale~$r$ & SI-$\delta_1$\,(\%) & Ratio~$\rho$ \\
\midrule
DA3~\cite{depthanything3}          & 1.4\,B  & 4.16  & 99.6 & 0.67 & 89.1 & 53.6 \\
MapAnything~\cite{keetha2026mapanything} & 1.1\,B  & 10.88 & 99.3 & \underline{0.75} & 80.7 & 156.1 \\
PPD~\cite{xu2025pixel}       & 831\,M  & 16.21 & 92.0 & 0.66 & \underline{91.3} & 90.4 \\
MoGe-2~\cite{wang2025moge2}         & 331\,M  & 16.10 & \underline{100.0} & 0.62 & 91.1 & 81.0 \\
\midrule
\DDS{}   & \textbf{2.1\,M\,ft} & \textbf{1.11}  & \underline{100.0} & 0.28 & 91.1 & \underline{9.2} \\
\textbf{DermDepth}     & \textbf{2.1\,M\,ft} & \underline{0.87} & \underline{100.0} & \textbf{0.91} & \textbf{92.6} & \textbf{1.95} \\
\bottomrule
\end{tabular}
\end{table}

\begin{figure}[t]
    \centering
    \includegraphics[width=\textwidth]{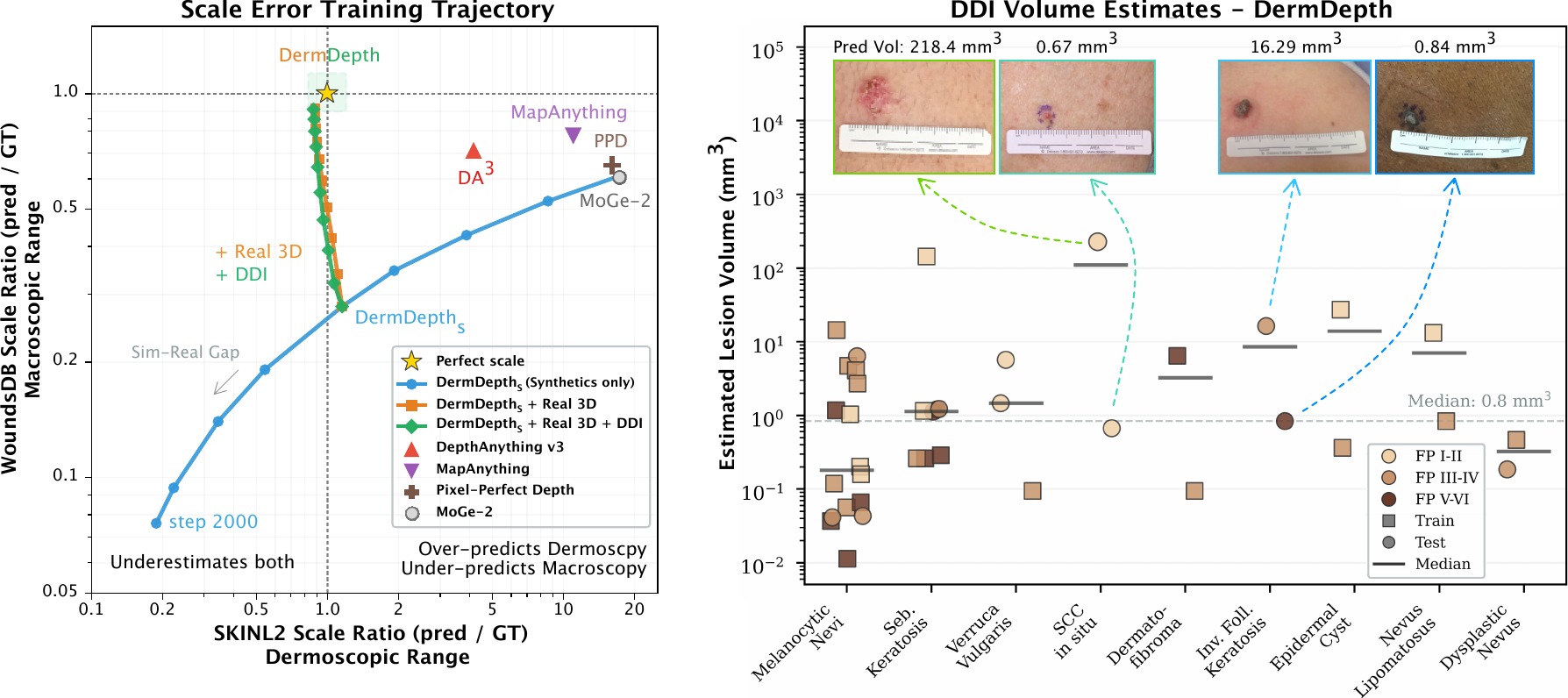}
    \caption{\textbf{Scale Correction Training (Left), Predicted Volume (Right).}
    We measure dermoscopic range (cm-level) over-estimation and macroscopic range (m-level) underestimation for all foundation models tested. Training on D-Synth corrects dermoscopic error, adding real data corrects macroscopic error (left). We demonstrate DermDepth volume measurements on DDI held-out test data (right).}
    \label{fig:quantitative}
\end{figure}

\begin{figure}[t]
    \centering
    \includegraphics[width=\textwidth]{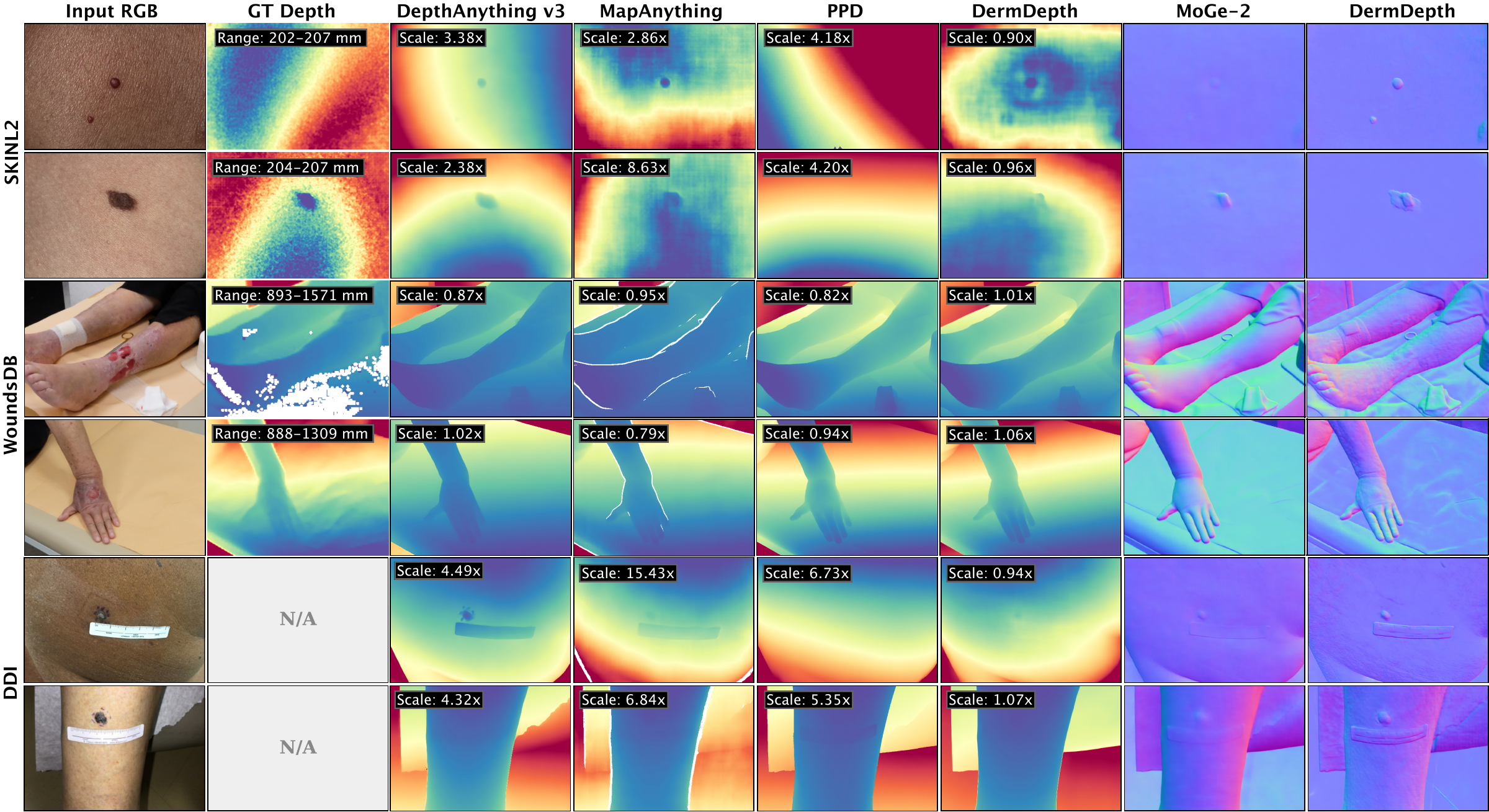}
    \caption{\textbf{Predicted Depth, Scale and Normals.}
    We visualize performance across three datasets at different metric distances. We find DermDepth produces more accurate and consistent scale and normal maps with finer lesion detail. We note DDI was not collected alongside 3D information, thus we estimate scale error from known ruler area. \new{Ground-truth may show dataset-specific sensor noise, as discussed in Sect.~\ref{sec:realdata}.}}
    \label{fig:comparison}
\end{figure}

\subsection{Baseline Methods and Metrics}
We benchmark four recent models that claim metric depth output:
\textbf{Depth Anything~3}~\cite{depthanything3} (DA3-METRIC-LARGE; 1.4\,B params),
\textbf{MapAnything}~\cite{keetha2026mapanything} (1.1\,B),
\textbf{Pixel-Perfect Depth}~\cite{xu2025pixel} (PPD; 831\,M, diffusion refinement with MoGe-2 scaling as done on the official codebase), and
\textbf{MoGe-2}~\cite{wang2025moge2} (331\,M).

We report three complementary metrics:
\textbf{Scale ratio} $r = \text{median}(\hat{d}_i / d_i^*)$, where $r{=}1.0$ is perfect;
\textbf{SI-$\delta_1$} = percentage of pixels where the scale-aligned depth ratio falls within $[1/1.25,\, 1.25]$, measuring geometry quality independent of global scale; and
\textbf{DDI ratio} $\rho = A_{\text{pred}} / A_{\text{GT}}$, measuring predicted ruler area relative to the known ruler area of 6.6\,cm$^2$.

\subsection{Experimental Results}
\label{sec:main_results}

\cref{tab:main} demonstrates that every evaluated foundation model produces systematically incorrect metric scale on dermatological images. All four baselines overestimate dermoscopic depth by 4--16$\times$ (SKINL2) and underestimate clinical wound depth at 0.6--0.75$\times$ (WoundsDB), with DDI ruler-area ratios ranging from 54$\times$ to 156$\times$. Crucially, geometry quality is high across all models (SI-$\delta_1 \geq 92\%$), confirming that the 3D structure is captured correctly, supporting our decision to freeze the majority of parameters, focusing on scale and normal map training.

PPD internally uses MoGe-2 for metric grounding and consequently inherits near-identical scale bias (16.2$\times \approx$ 16.1$\times$ on SKINL2), while its diffusion refinement slightly degrades geometry (SI-$\delta_1$: 92.0\% vs.\ 100.0\%). DA3, despite having the most parameters, overestimates by 4.2$\times$.

DermDepth corrects scale across all three benchmarks. With synthetic data alone, SKINL2 scale drops from 16.1$\times$ to 1.11$\times$ while high-quality geometry is preserved (SI-$\delta_1${=}100\% and SI-AbsRel{=}0.017). With progressive real-data training, DermDepth achieves 0.87$\times$ on SKINL2, 0.91$\times$ on WoundsDB, and 1.95$\times$ on DDI, bringing all three benchmarks close to metric scale, as shown in \cref{fig:quantitative} (left).

\subsection{\new{Training Strategy Ablations}}\label{sec:ablation}

\new{In the small-data-large-model setting it is common to freeze most model weights and train only output heads; we ablate this choice. Under synthetic-only training, unfreezing more parameters ($2.1$--$326$\,M) progressively harms scale and geometry on real data: $3{,}000$ synthetic samples cannot re-learn the full decoder, and the sim-real gap causes overfitting (\cref{tab:ablation}, rows~1--4). D-Synth covers only dermoscopic distances ($\sim12$--$20$\,mm); adding real SKINL2 ($\sim$few-hundred mm) and WoundsDB ($\sim1$--$1.5$\,m) depth bridges the macroscopic gap but worsens DDI, which sits between the two in sampling distance. Including $33$ DDI pseudo-GT samples anchored to ruler measurements generalizes scale prediction, without compromising other benchmarks (\cref{tab:ablation}, rows~2,~5--6), justifying our output-head-only and progressive real data training design.}



\begin{table}[t]
\centering
\caption{\new{\textbf{Training ablations, trainable parameters and real-data value.}}}
\label{tab:ablation}
\small
\setlength\tabcolsep{4pt}
\new{\begin{tabular}{@{}l ccc cc@{}}
\toprule
 & \multicolumn{3}{c}{Scale ratio $r$ ($\to 1.0$)} & \multicolumn{2}{c}{SI-$\delta_1$ (\%)} \\
\cmidrule(lr){2-4} \cmidrule(lr){5-6}
Configuration & SKINL2 & WoundsDB & DDI & SKINL2 & WoundsDB \\
\midrule
\hspace{0.75em}Baseline (MoGe-2)                 & 16.10 & 0.62 & 81.0 & 100.0 & 91.1 \\
\rowcolor{green!6}
\hspace{0.75em}Scale head, synth                  & \textbf{1.11}  & 0.28 &  9.2 & 100.0 & 91.1 \\
\hspace{0.75em}+ Decoder, synth                   & 0.14  & 0.06 & 0.76 &  99.9 & 85.8 \\
\hspace{0.75em}+ Full model, synth                & 0.09  & 0.04 & 0.13 &  99.8 & 77.5 \\
\midrule
\hspace{0.75em}Scale head, +\,real depth          & \textbf{0.89}  & \textbf{0.92} & 18.3 & 100.0 & 91.1 \\
\rowcolor{green!6}
\hspace{0.75em}\textbf{Scale head, +\,DDI pseudo} & \underline{0.87}  & \underline{0.91} & \textbf{1.95} & \textbf{100.0} & \textbf{92.6} \\
\bottomrule
\end{tabular}}
\end{table}

\subsection{Skin Tone Fairness}
\label{sec:fairness}

Equitable performance across skin tones is essential for clinical deployment of dermatology methods~\cite{daneshjou_2022_disparities}.
\cref{tab:fairness} evaluates DDI ruler-area ratio stratified by Fitzpatrick group. MapAnything shows the strongest bias (211.88$\times$ on FP~V--VI vs.\ 88.53$\times$ on FP~III--IV), while the most uniform baseline, MoGe-2, has a spread of 10.90. Diverse skin tones in our synthetic training data could be valuable to reduce bias as \DDS{} lowers disparity to 1.70, while training on real data leads to 1.02, with per-group ratios between 1.39$\times$ and 2.41$\times$.

\begin{table}[t]
\centering
\caption{\textbf{DDI ruler-area ratio by Fitzpatrick skin tone.}
DermDepth achieves equitable performance across skin tones, reducing the disparity (max\,$-$\,min per-group ratio). \DDS{}: synthetic data only; DermDepth: synth + SKINL2, DDI, WDB.
}
\label{tab:fairness}
\small
\setlength\tabcolsep{4pt}
\begin{tabular}{@{}l cccc c@{}}
\toprule
Method & Overall & FP\,I--II & FP\,III--IV & FP\,V--VI & Disparity \\
\midrule
DA3~\cite{depthanything3}               & 53.60$\times$ &  73.00$\times$ &  58.60$\times$ &  38.10$\times$ & 34.90 \\
MapAnything~\cite{keetha2026mapanything} & 156.11$\times$ & 156.11$\times$ &  88.53$\times$ & 211.88$\times$ & 123.35 \\
PPD~\cite{xu2025pixel}                  & 90.36$\times$ & 100.44$\times$ & 100.10$\times$ &  80.26$\times$ & 20.18 \\
MoGe-2~\cite{wang2025moge2}             & 81.00$\times$ &  88.50$\times$ &  84.50$\times$ &  77.60$\times$ & 10.90 \\
\midrule
\DDS{}                 &  9.20$\times$ &  9.20$\times$ & 10.00$\times$ &  8.30$\times$ &  1.70 \\
\textbf{DermDepth}              & \textbf{1.95$\times$} & \textbf{1.92$\times$} & \textbf{2.41$\times$} & \textbf{1.39$\times$} & \textbf{1.02} \\
\bottomrule
\end{tabular}
\end{table}

\subsection{Normal Map Qualitative Results}
\label{sec:normals}

Surface normal maps encode local 3D orientation at each pixel, capturing micro-geometric features invisible in RGB or depth alone. As shown in \cref{fig:comparison}, DermDepth produces normal maps with noticeably finer lesion detail compared to the baseline predictions. The improved normals better delineate lesion boundaries, surface roughness and elevation changes, features directly relevant to dermatological assessment. In particular, the Asymmetry and Border components of ABCDE screening~\cite{blum-2003} and texture features such as ulceration, scaling and papillomatous morphology manifest as distinctive normal map patterns. Quantitative normal descriptors derived from these maps could supplement existing 2D analysis by capturing surface morphology currently assessable only through physical examination or specialized hardware.

\subsection{Clinical Use Case: 3D Lesion Measurement}
\label{sec:measurement}

We validate clinical utility by measuring lesion geometry in DDI images with known-sized rulers. We highlight test-set cases of Squamous Cell Carcinoma and Inverted Follicular Keratosis to visually validate reasonable volumetric differences (\cref{fig:quantitative}, right). To verify if DermDepth width estimates are clinically plausible, we collect dermatological literature that has recorded observations of test-set DDI lesions. Across 8 diagnostic categories (14 lesions), 10 of 14 fall within published typical ranges, while the remaining 4 could be clinically viable boundary cases: an early-stage keratoacanthoma (7.1\,mm vs.\ classic mature 10--20\,mm), an early-detection Bowen's lesion (4.5\,mm vs.\ median 10\,mm), a dysplastic nevus at the $\geq$5\,mm diagnostic threshold (5.1\,mm), and a large melanocytic nevus in the top 2\% by size (9.0\,mm). These ranges along with citations are shown in \cref{tab:litval}.

\begin{table}[t]
\centering
\caption{\textbf{Literature validation of DermDepth lesion width predictions.}
Predicted widths for DDI test-set lesions compared to published clinical size ranges.
}
\label{tab:litval}
\small
\setlength\tabcolsep{4pt}
\begin{tabular}{@{}l c l c@{}}
\toprule
Lesion type & DermDepth (mm) & Published range (mm) & Sources \\
\midrule
Seborrheic keratosis      & 5.9--10.0 & 1--30; typical 5--10           & \cite{greco-2024,barthelmann-2023} \\
Melanocytic nevi           & 1.6--9.0  & 1--10; 98\% $\leq$5           & \cite{iyidal-2016,muradia-2022} \\
Inv.\ follicular keratosis & 9.2--10.2 & 3--30; mean 10.1  & \cite{hamzelou-2025} \\
Verruca vulgaris           & 7.2--7.4  & 2--10 & \cite{aboud-2023} \\
SCC, keratoacanthoma       & 7.1       & 10--20 (mature) & \cite{zito-2023} \\
Bowen's disease            & 4.5       & 1--55; median 10 & \cite{fougelberg-2021} \\
Melanoma                   & 9.9       & IQR 9--20 & \cite{dessinioti-2024} \\
Dysplastic nevus           & 5.1       & $\geq$5; typical 5--15 & \cite{baigrie-2022} \\
\bottomrule
\end{tabular}
\end{table}

\section{Discussion and Conclusion}
\label{sec:discussion}

\paragraph{Limitations.}
D-Synth currently renders dermoscopic-distance scenes (12--20\,mm), so macroscopic clinical distances must be covered by real training data, a gap we bridge with WoundsDB and SKINL2, but which limits the synthetic component to close-range correction.
The real-data ground-truth itself carries noise. These issues affect geometric evaluation (surface normals in particular) but do not compromise metric scale. Finally, DDI pseudo-GT is derived from ruler segmentation and known ruler dimensions; segmentation errors or partial ruler occlusion introduce uncertainty, and the DDI test set ($n{=}14$) remains small, limiting statistical significance for DDI results, an issue common under limited fairness data. We strongly believe the community should collect more data with clean 3D information.

\paragraph{Conclusion.}
We have shown that foundation metric depth models fail on dermatological images, overestimating dermoscopic depth by 4--16$\times$ while underestimating clinical wound depth.
DermDepth corrects this systematic error by fine-tuning a scale prediction head and normal output head (0.6\% of parameters) on D-Synth, the first synthetic dermoscopic dataset with pixel-perfect 3D information, followed by progressive real-data refinement with just 346 clinical samples. The result is much improved metric scale across three diverse benchmarks with equitable performance across Fitzpatrick skin tones (disparity: 10.90~$\rightarrow$~1.02). \new{DermDepth is memory efficient, and could run on mobile devices.} We publish all code, models, and datasets to encourage further research in this direction.


\begin{credits}
\subsubsection{\discintname}
\new{The authors have no competing interests to declare that are relevant to the content of this article.}
\end{credits}

\bibliographystyle{splncs04}
\bibliography{main}

\end{document}